\documentclass{article} 
\usepackage{iclr2026_conference,times}

\usepackage{hyperref}
\usepackage{url}
\usepackage{enumitem}
\usepackage{multirow}
\usepackage{booktabs}
\usepackage{graphicx}
\usepackage{amsmath}
\usepackage{amsfonts}

\title{In-Context Learning with Unpaired Clips for Instruction-based Video Editing}

\author{
  \textbf{Xinyao Liao}$^{1,2}$\quad
  \textbf{Xianfang Zeng}$^{2}$\footnotemark[1]\quad
  \textbf{Ziye Song}$^{1}$\quad
  \textbf{Zhoujie Fu}$^{1,2}$\quad
  \textbf{Gang Yu}$^{2}$\footnotemark[2]\quad
  \textbf{Guosheng Lin}$^{1}$\footnotemark[2]\\[1ex]
  $^{1}$Nanyang Technological University \quad
  $^{2}$StepFun\\[1ex]
}

\iclrfinalcopy 
\begin{document}

\maketitle
\renewcommand{\thefootnote}{\fnsymbol{footnote}}
\footnotetext[1]{Xianfang Zeng is the project leader.}
\footnotetext[2]{Corresponding authors: skicy@outlook.com, gslin@ntu.edu.sg}
\renewcommand{\thefootnote}{\arabic{footnote}}
\begin{abstract}
Despite the rapid progress of instruction-based image editing, its extension to video remains underexplored, primarily due to the prohibitive cost and complexity of constructing large-scale paired video editing datasets.
To address this challenge, we introduce a low-cost pretraining strategy for instruction-based video editing that leverages in-context learning from unpaired video clips. We show that pretraining a foundation video generation model with this strategy endows it with general editing capabilities, such as adding, replacing, or deleting operations, according to input editing instructions. The pretrained model can then be efficiently refined with a small amount of high-quality paired editing data.
Built upon HunyuanVideoT2V, our framework first pretrains on approximately 1M real video clips to learn basic editing concepts, and subsequently fine-tunes on fewer than 150k curated editing pairs to extend more editing tasks and improve the editing quality. Comparative experiments show that our method surpasses existing instruction-based video editing approaches in both instruction alignment and visual fidelity, achieving a 12\% improvement in editing instruction following and a 15\% improvement in editing quality. \url{https://github.com/leoisufa/ICVE}
\end{abstract}    
\section{Introduction}
Nowadays, the field of Artificial Intelligence Generated Content (AIGC) is no longer restricted to text-based control of generated results. Various conditioning methods are introduced to enable more fine-grained regulation of content generation \citep{ma2025controllable}. Beyond accurately producing desired outputs under different control conditions, the editing of existing images and videos also emerges as an important research direction.

To accomplish video editing, various approaches are proposed. Training-free methods modify video content by operating on attention maps in a sophisticated manner without retraining the foundation model \citep{qi2023fatezero,cong2023flatten,kara2024rave,geyer2023tokenflow,li2024vidtome,ku2024anyv2v,fan2024videoshop,wang2025videodirector,li2025flowdirector}. Mask-based methods employ masks as input conditions to explicitly specify editing regions, thereby providing precise guidance for both the editing location and content \citep{hu2024vivid,jiang2025vace,gao2025lora}. However, these approaches exhibit limitations: training-free methods often compromise generation quality, while mask-based methods require additional steps to obtain segmentation results of the targeted regions. Recently, instruction-based editing models emerge. These methods require only textual input and the original video. In image editing, instruction-based approaches already achieve rapid progress. Representative examples include closed-source models such as ChatGPT \citep{GPT-5} and Gemini \citep{Gemini-2.5}, as well as open-source models such as Qwen-Image-Edit \citep{wu2025qwen}, Step1X-Edit \citep{liu2025step1x}, and FLUX.1 Kontext Dev \citep{labs2025flux}. In contrast, in video editing, a high-quality instruction-based model remains underexplored.

A major challenge in this task is that training a high-quality model requires millions of paired editing samples. Compared with image editing datasets, video editing data are scarcer in source and more difficult to process. Real editing data can be obtained through direct shooting or rendering, but the scale of real data remains limited. Alternatively, training data can be distilled from existing editing models \citep{cheng2023consistent,wu2025insvie,zi2025se}, yet synthetic datasets often introduce artifacts and require substantial computational resources. The shortage of high-quality training data therefore constitutes a critical obstacle, highlighting the need for approaches that reduce dependence on large-scale annotated video editing datasets.

To address the data challenge and achieve a high-quality model, we propose a two-stage training strategy and modify an in-context video editing framework. The model is first pretrained on video clip data and then fine-tuned on a small number of editing pairs. Specifically, a long video segment without transitions is divided into multiple short clips, from which two clips are randomly selected as the original and the pseudo-edited video. An editing instruction is then generated based on their differences, and these video clips are used in the pretraining stage to teach the model basic editing concepts. In the subsequent supervised fine-tuning (SFT) stage, a small amount of synthetic editing data is used to enhance the model’s editing capability further. For the model architecture, we employ in-context input, where the tokens of the original video are concatenated with the noised tokens. To adapt to the noiseless property of the original video, the timesteps corresponding to its tokens are set to 0. Overall, with this strategy, our model relies on about 1M video clips and fewer than 150k multi-task editing pairs, yet achieves superior performance compared with models trained on several million editing data.

In the experiments, we first compare our model with existing instruction-based video editing methods. The comparisons are evaluated along two dimensions: instruction-following capability and visual quality of the generated videos. In both aspects, our model achieves state-of-the-art performance. In the ablation study, we assess different training strategies, and the results show that the proposed approach allows the model to acquire basic editing capabilities during video clip pretraining and to reach superior performance after SFT. The contributions are summarized as follows:
\begin{itemize}[left=0pt]
\item A two-stage training strategy is proposed, consisting of video clip pretraining for basic editing capabilities generalization and SFT on editing data to extend editing types and improve the quality of generated video. 
\item An instruction-based video editing model is developed, which achieves superior performance over existing video editing approaches with approximately 12\% improvement in the editing instruction following and 15\% in the editing quality. 
\end{itemize}
\section{Related Work}
\textbf{Training-free Video Editing.}
Training-free video editing methods are primarily categorized into two paradigms. The first is the inversion-based approach, which follows an invert-then-denoise pipeline. FateZero \citep{qi2023fatezero} constrains the randomness of diffusion models to enable consistent editing. FLATTEN \citep{cong2023flatten} integrates optical flow into the attention module to improve visual consistency. RAVE \citep{kara2024rave} employs a noise-shuffling strategy to achieve temporally consistent and memory-efficient editing. TokenFlow \citep{geyer2023tokenflow} propagates intermediate features across frames to enforce temporal consistency while preserving spatial layout and motion. VidToMe \citep{li2024vidtome} applies a token-merging strategy that compresses self-attention tokens for temporal coherence. AnyV2V \citep{ku2024anyv2v} leverages the edited first frame and the features of an image-to-video (I2V) model to guide the edited sequence. Videoshop \citep{fan2024videoshop} adopts the modified first frame and propagates the prior through latent inversion with noise extrapolation. VideoDirector \citep{wang2025videodirector} introduces spatial-temporal decoupled guidance and multi-frame null-text optimization to enhance pivotal inversion. However, inversion approximation errors substantially degrade editing fidelity. An alternative is the inversion-free paradigm, which directly maps an Ordinary Differential Equation (ODE) path between the original and edited distributions. FlowEdit \citep{kulikov2024flowedit} proposes an inversion-free, optimization-free, and model-agnostic text-based editing framework, which is further applied to HunyuanLoom \citep{HunyuanLoom} for video editing. FlowDirector \citep{li2025flowdirector} models editing as an ODE-guided evolution in data space, enhanced with guidance strategies to improve semantic alignment. Although these training-free methods exploit the prior knowledge of foundation video generative models without additional training, their performance remains limited due to the inherent gap between generation and editing tasks.

\textbf{Training-based Video Editing.}
To overcome the limitations of training-free methods, several training-based approaches have been introduced. Tune-A-Video \citep{wu2023tune} presents a one-shot video tuning framework that adapts a text-to-image model with spatio-temporal attention for single-video editing. Video-P2P \citep{liu2024video} tunes a text-to-set model through an approximate inversion strategy. These one-shot methods improve temporal consistency, but the optimization process for each video remains time-consuming. For more general video editing, mask-based methods are developed, supported by the construction of large-scale mask-based video editing datasets. VIVID-10M \citep{hu2024vivid} introduces a hybrid image–video dataset and proposes a versatile interactive model with keyframe-guided propagation. VACE \citep{jiang2025vace} builds on Wan Video \citep{wan2025wan} and supports multiple input conditions through Context Adapter Tuning. LoRA-Edit \citep{gao2025lora} proposes a mask-based LoRA tuning approach for region-specific video editing. To provide more convenient editing, instruction-based methods have been proposed. InstructVid2Vid \citep{qin2024instructvid2vid} introduces a data pipeline for generating paired training data. InsV2V \citep{cheng2023consistent} constructs a dataset with more than 400K synthetic editing pairs and trains an editing model on it. EffiVED \citep{zhang2024effived} proposes an automatic pipeline for constructing editing data from image datasets and real-world source videos. FlowV2V \citep{wang2025consistent} reformulates video editing as flow-driven I2V generation by combining first-frame editing with pseudo flow simulation. Lucy Edit \citep{Lucy_Edit} concatenates the original video with noise along the channel dimension, thereby reducing the token context length. InstructVEdit \citep{zhang2025instructvedit} introduces a full-cycle instruction-based video editing approach. InsViE-1M \citep{wu2025insvie} provides a large-scale, high-quality dataset with 1M editing pairs and develops a multi-task video editing model. Señorita-2M \citep{zi2025se} constructs a dataset of 2M high-quality editing pairs using state-of-the-art specialized editing models. Most of these works rely on distilling existing video editing models or introducing additional control conditions to construct synthetic data. However, training exclusively on synthetic data inevitably introduces artifacts and an ``AI vibe.'' Moreover, generating large-scale synthetic datasets requires substantial computational resources for both inference and data cleaning.
\section{Method}
\subsection{Data Curation}
\begin{figure}[t]
\centering
\vspace{-0.8em}
\includegraphics[width=0.99\linewidth]{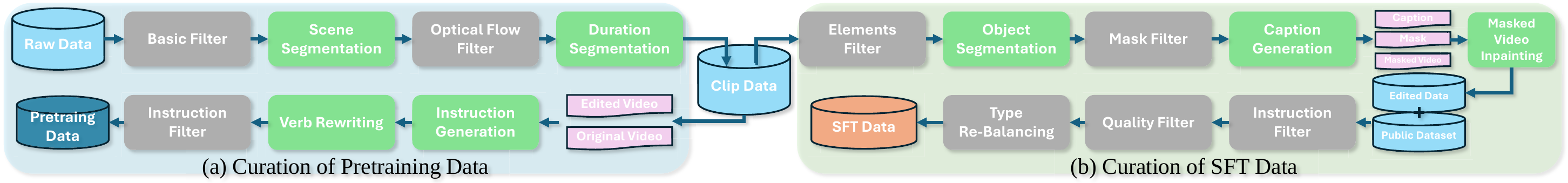}
\vspace{-0.8em}
\caption{Pipeline of training data curation. (a) shows the pipeline for curating clip data from raw videos; (b) illustrates the pipeline for synthesizing and filtering editing data.}
\vspace{-0.6em}
\label{fig:data}
\end{figure}
\textbf{Pretraining Clip Data.}  
Our approach is inspired by the data pre-processing pipelines employed in foundation video generation models \citep{kong2024hunyuanvideo,wan2025wan,gao2025seedance}. In these pipelines, raw videos are first segmented using a scene detection method \citep{SceneDetect}, producing multiple scene-level segments. After ensuring that no transitions appear within each segment, these segments are further divided into clips of fixed duration according to the target number of frames required by the foundation model. In the context of video editing, an original video is provided as input, with specific parts modified while surrounding scenes, characters, and objects are preserved. Similarly, clips extracted from the same scene generally share highly similar visual content, such as the same backgrounds, characters, and objects. Since they are sampled from different temporal intervals, variations naturally occur, including camera motion, changes in character positions, and object movement. By treating one clip as the original video and another as the pseudo-edited video, the pair can be used as a training sample, where the transformation from the original to the pseudo-edited clip is regarded as a form of video editing.  

Building on this idea, we adapt the data curation pipeline of foundation models for editing data collection. As illustrated in Figure \ref{fig:data}(a), after applying basic filters on duration, resolution, and frame rate, raw videos are divided into scene-level segments. Optical flow is then computed for each segment to filter out those with low motion amplitude. The remaining segments are subsequently divided into non-overlapping clips along the timeline according to the required duration. From each segment, two clips are randomly selected as the original and pseudo-edited videos and are annotated by Step3 \citep{wang2025step}. The annotation process generates an instruction that describes the transformation from the original clip to the pseudo-edited one. To further enrich the data, two forms of augmentation are applied:  
(1) rewriting action verbs in the editing instructions, such as replacing ``replace'' with synonyms like ``change'' or ``modify''; and  
(2) filtering out trivial instructions that involve meaningless modifications such as ``brightness,'' ``contrast,'' or ``saturation.'' Through this pipeline, large-scale video clip pairs are curated for pretraining.

\textbf{SFT Editing Data.}  
Publicly available video editing training data remain scarce, particularly with the emergence of advanced foundation models that require high resolution, high frame counts, and high frame rates. The quality of existing datasets often fails to meet the requirements of these latest video models. To construct high-quality editing data, we design a synthetic data pipeline.

Specifically, our pipeline employs VACE \citep{jiang2025vace} as the basic video inpainting model. As illustrated in Figure \ref{fig:data}(b), video clips are first filtered using Step3 \citep{wang2025step}, which labels persons and objects in the foreground as well as elements in the background. Objects that are difficult to describe or unsuitable for editing are removed. GroundedSAM2 \citep{ren2024grounded,ravi2024sam} is then used to obtain segmentation masks. Based on these masks, a threshold is applied to exclude unsuitable targets. In particular, edited instances are required not to exceed $50\%$ of the frame area in at least $80\%$ of the frames. This constraint prevents large discrepancies between the edited and original videos, thereby ensuring stable training. Next, the mask boundaries are randomly expanded by 20–50 pixels to prevent the inpainted regions from maintaining identical shapes and sizes to the original instances. The masked regions in the original video are then set to black, and Step3 generates a caption based on the unmasked regions. Finally, the mask, the masked video, and the caption are fed into VACE \citep{jiang2025vace} to perform inpainting, producing the edited video. The original and edited videos are then annotated again by Step3 to generate an editing instruction.

After the raw editing data are generated, a multi-stage filtering process is applied to improve quality. First, videos containing large black silhouette regions, which indicate inpainting failure, are discarded. Second, Qwen2.5-VL \citep{bai2025qwen2} evaluates the remaining training data along two dimensions: the accuracy of the editing instruction and the visual quality of the edited video, each scored on a 1–5 scale. Since VACE does not generate style editing data, style-related samples are additionally collected from InsViE-1M \citep{wu2025insvie} and Señorita-2M \citep{zi2025se}. For the SFT stage, only samples with a score of 5 in both dimensions are retained. Finally, the amount of data from different editing types is balanced to avoid bias.

\begin{figure}[t]
\centering
\vspace{-0.8em}
\includegraphics[width=0.99\linewidth]{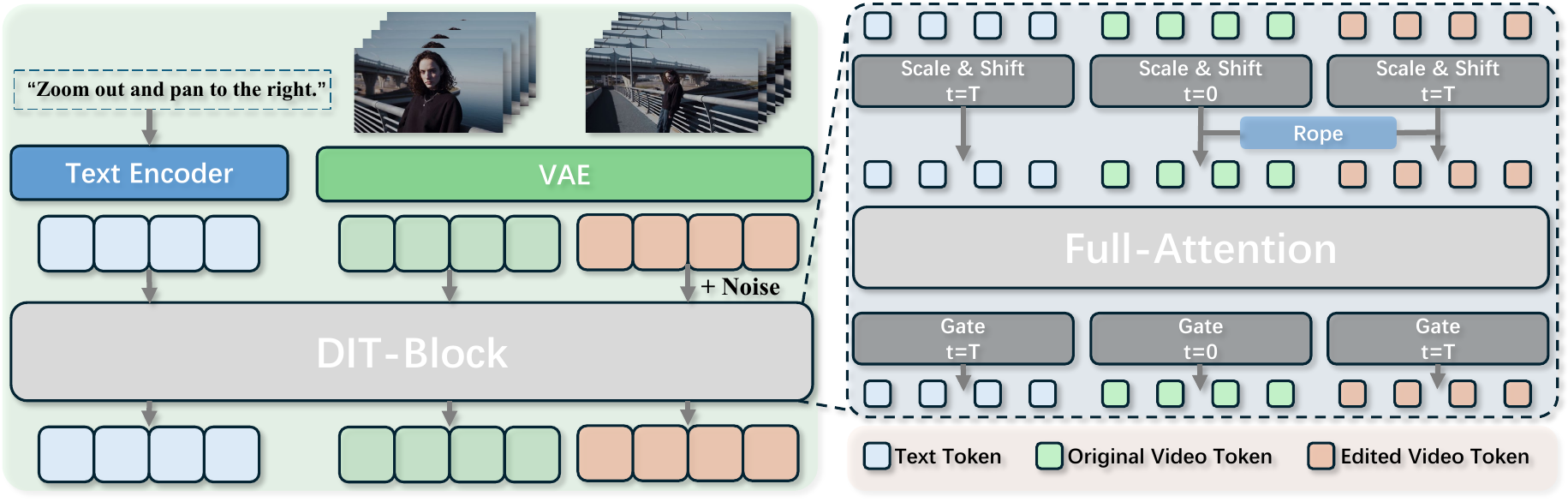}
\vspace{-0.8em}
\caption{In-context Instruction-based Video Editing Model. The instruction, original video, and edited video are injected into the model in an in-context manner. The timesteps corresponding to the original video tokens are fixed to 0, while the timesteps of text and edited video tokens retain T.}
\vspace{-0.8em}
\label{fig:model}
\end{figure}
\subsection{In-Context Instruction-based Video Editing Model}

The proposed method adopts HunyuanVideoT2V \citep{kong2024hunyuanvideo} as the foundation video generation model. To adapt the base model to the instruction-based video editing task, several structural modifications are introduced. Unlike text-to-video (T2V) generation, the input prompts here are editing instructions rather than textual descriptions of video content. Therefore, the prompt prefixes used in the text encoder for enriching detailed descriptions are removed, and the editing instructions are directly fed into the text encoder without additional prefixes.  

In addition to the instruction prompt, the original video is also required as input. As illustrated in Figure~\ref{fig:model}, the original video is injected into the network in an in-context manner. Specifically, in the model, both the original and the noised videos are converted from latents to tokens using the same $2\times2\times1$ patchify module. The tokens of the original video and the noised video are then concatenated along the sequence dimension, formulated as:
\begin{equation}
    x_{\text{orig, noise}}^{t} = \text{Cat}[\textbf{P}(z_{\text{orig}}), \textbf{P}(z_{\text{noise}}^{t})],
\end{equation}
where $z_{\text{orig}}$ and $z_{\text{noise}}^{t}$ denote the pure latents of the original video and the noised latents of the editing video, respectively. $\textbf{P}$ is the patchify module, and $\text{Cat}$ denotes concatenation along the sequence dimension. $x_{\text{orig, noise}}^{t}$ represents the total visual tokens for the DiT block input.  

Since the original video is provided without added noise, the timesteps corresponding to the original video tokens are fixed to 0, simulating the noise-free case. The text tokens and noised tokens retain their original timesteps in the range of 0–1. These distinct timesteps affect the modulation operation in the DiT block, which predicts different scale, shift, and gate parameters according to the input timesteps. This process is formulated as:
\begin{equation}
    x_{\text{text, orig, noise}}^{t} = \text{Cat}[\textbf{M}(x_{\text{text}}, t=T), \textbf{M}(x_{\text{orig}}, t=0), \textbf{M}(x_{\text{noise}}^{t}, t=T)],
\end{equation}
where $\textbf{M}$ denotes the modulation function, including scale, shift, and gate. $x_{\text{text}}$, $x_{\text{orig}}$, and $x_{\text{noise}}^{t}$ represent the text tokens, original video tokens, and noised edited video tokens, respectively, and $t$ indicates the input timesteps.  

With this design, the DiT block treats the original video tokens as noise-free tokens with a timestep of 0, thereby preserving the visual information of the original video and ensuring high-quality generation. Moreover, by applying distinct scale, shift, and gate parameters to the original and noised tokens, the distributions of the two token types remain separated in the attention calculation, enabling the model to distinguish between them effectively. After these modifications, our model directly takes the original video and the instruction prompt as inputs. The noised tokens inherit visual content from the original video and modify specific elements under the guidance of the instruction prompt, thereby accomplishing instruction-based video editing.  

\subsection{Training Details}

\textbf{Objective Function.}  
During training, the editing instruction prompt $y$, the original video $V_{\text{orig}}$, and the edited video $V_{\text{edit}}$ are provided. Both videos are first encoded by the VAE $\mathcal{E}$ to produce video latents $z_{\text{orig}} = \mathcal{E}(V_{\text{orig}})$ and $z_{\text{edit}} = \mathcal{E}(V_{\text{edit}})$. The editing instruction is fed into the text encoder to generate text tokens $x_{\text{text}}$. Noise $\epsilon$ is added to the edited latent $z_{\text{edit}}$ to obtain the noised latent $z_{\text{noise}}^{t}$ over timesteps $t \in (0,1)$. The video editing model then predicts the noise $\epsilon$ added to $z_{\text{edit}}$, conditioned on $t$, $z_{\text{orig}}$, and $x_{\text{text}}$. The training objective is defined as:
\begin{equation}
    \mathcal{L} = \mathbb{E}_{\epsilon \sim \mathcal{N}(0,1),\, t} 
    \Bigl[ \bigl\| \epsilon - \epsilon_\theta(t, z_{\text{noise}}^{t}, z_{\text{orig}}, x_{\text{text}}) \bigr\|_2^2 \Bigr],
\end{equation}
where $\epsilon_\theta$ denotes the noise predicted by the video editing model parameterized by $\theta$.  

\textbf{Pretraining via Clip Data.}  
In the pretraining stage, the training process relies solely on video clips without using any editing data. Pretraining on clips offers several advantages. First, existing data pre-processing pipelines from foundation video models \citep{wan2025wan,kong2024hunyuanvideo,gao2025seedance} can be directly reused for data collection. These pipelines are simple and mature, enabling the acquisition of large volumes of usable data. Second, video clips from real-world sources generally exhibit higher visual quality than synthetic editing data, as they are free from artifacts, ``AI vibe,'' or other implausible elements. Video clips extracted from the same scene segment help the model learn to preserve contextual information of the original video, including scene layout, character identity, and object appearance. This strengthens the model’s ability to retain original video content during editing operations. Meanwhile, clips sampled from different temporal intervals introduce motion variations, which can be regarded as a form of video-to-video editing. Although such clip data are not strictly aligned editing pairs along the same timeline, they still allow the model to learn basic editing concepts from temporal differences. Even with pretraining only on video clips, the model acquires initial editing capabilities, as later demonstrated in the ablation study.  

Specifically, pretraining begins at 240p resolution, leveraging large-scale video clips to guide the model in extracting and preserving scene, character, and object information from the original video input. At this stage, the model also develops a basic ability to follow simple editing instructions, such as addition, removal, and replacement operations. The resolution is then progressively increased from 240p to multiple resolution buckets to further enhance visual quality. Overall, this pretraining stage consumes approximately 1M video clip data.

\textbf{SFT on Editing Data.}  
After pretraining, the model is capable of preserving the content of the original video and demonstrates basic editing abilities. However, clip data does not provide strict editing concepts along the timeline and cannot represent stylized editing types. Therefore, SFT on high-quality editing data is required.  

Following prior strategies for foundation models \citep{wan2025wan,kong2024hunyuanvideo,gao2025seedance}, it is observed that SFT requires only a small amount of data and limited training time. Our SFT process adopts fewer than 150k high-quality editing video pairs and is conducted for one epoch. This setup strengthens the model’s editing capabilities while avoiding collapse and reducing the risk of overfitting caused by the relatively small dataset size. After SFT, the model learns to respond precisely to editing instructions and to generate high-quality editing results. It also enhances the ability to preserve fine-grained details in regions that remain unchanged. Moreover, the model adapts from supporting limited editing types after pretraining to handling a wide range of editing tasks after SFT. In summary, approximately $75\%$ of the training is dedicated to pretraining on video clip data, while less than $25\%$ involves high-quality editing data for SFT.
\section{Experiments}
\subsection{Experimental Settings}
\textbf{Compared Methods.}  
We compare our model with both training-free and training-based video editing methods that accept instruction input. The training-free methods include AnyV2V \citep{ku2024anyv2v} and Videoshop \citep{fan2024videoshop}, while the training-based methods include InsV2V \citep{cheng2023consistent}, Señorita-2M \citep{zi2025se}, InsViE-1M \citep{wu2025insvie}, and Lucy Edit Dev \citep{Lucy_Edit}. Since AnyV2V, Videoshop, and Señorita-2M require the first edited frame as an additional input, results are reported with the aid of two instruction-based image editing models: InstructPix2Pix (+InsP2P) \citep{brooks2023instructpix2pix} and Qwen-Image-Edit (+Qwen) \citep{wu2025qwen}.  

\textbf{Testing Dataset.}  
Following previous works \citep{fan2024videoshop,zi2025se,wu2025insvie}, we construct a test set containing 300 video samples from the DAVIS \citep{pont20172017} and YouTubeVOS \citep{xu2018youtube} datasets, as well as videos from the Pexels website \citep{Pexels}. GPT-5 \citep{GPT-5} is employed to generate diverse editing instructions for these original videos.  

\textbf{Evaluation Metrics.}  
Evaluation is conducted from two perspectives: instruction following and video generation quality. For instruction following, CLIP text–image embedding similarity \citep{radford2021learning} and the Pick score \citep{kirstain2023pick} are employed. Since these metrics cannot fully capture editing effectiveness, GPT-5 \citep{GPT-5} is additionally adopted for automated evaluation. GPT-5 assesses each test sample along four dimensions: alignment between the instruction and the edited video (Instruction Following, \textbf{I\_F}), preservation of unedited regions from the original video (Original Video Preservation, \textbf{O\_P}), overall editing quality (Editing Quality, \textbf{E\_Q}), and success ratio of edits (Success Ratio, \textbf{S\_R}). Scores range from 1 to 5, and the reported results are averaged across all test samples. For video generation quality, we adopt the evaluation metrics from VBench \citep{huang2024vbench}. The generated videos are evaluated on Subject Consistency (\textbf{S\_C}), Background Consistency (\textbf{B\_C}), Motion Smoothness (\textbf{M\_S}), and Temporal Flickering (\textbf{T\_F}).  
\subsection{Comparison with SOTA video editing models}
\begin{figure}[ht]
\centering
\vspace{-0.8em}
\includegraphics[width=0.95\linewidth]{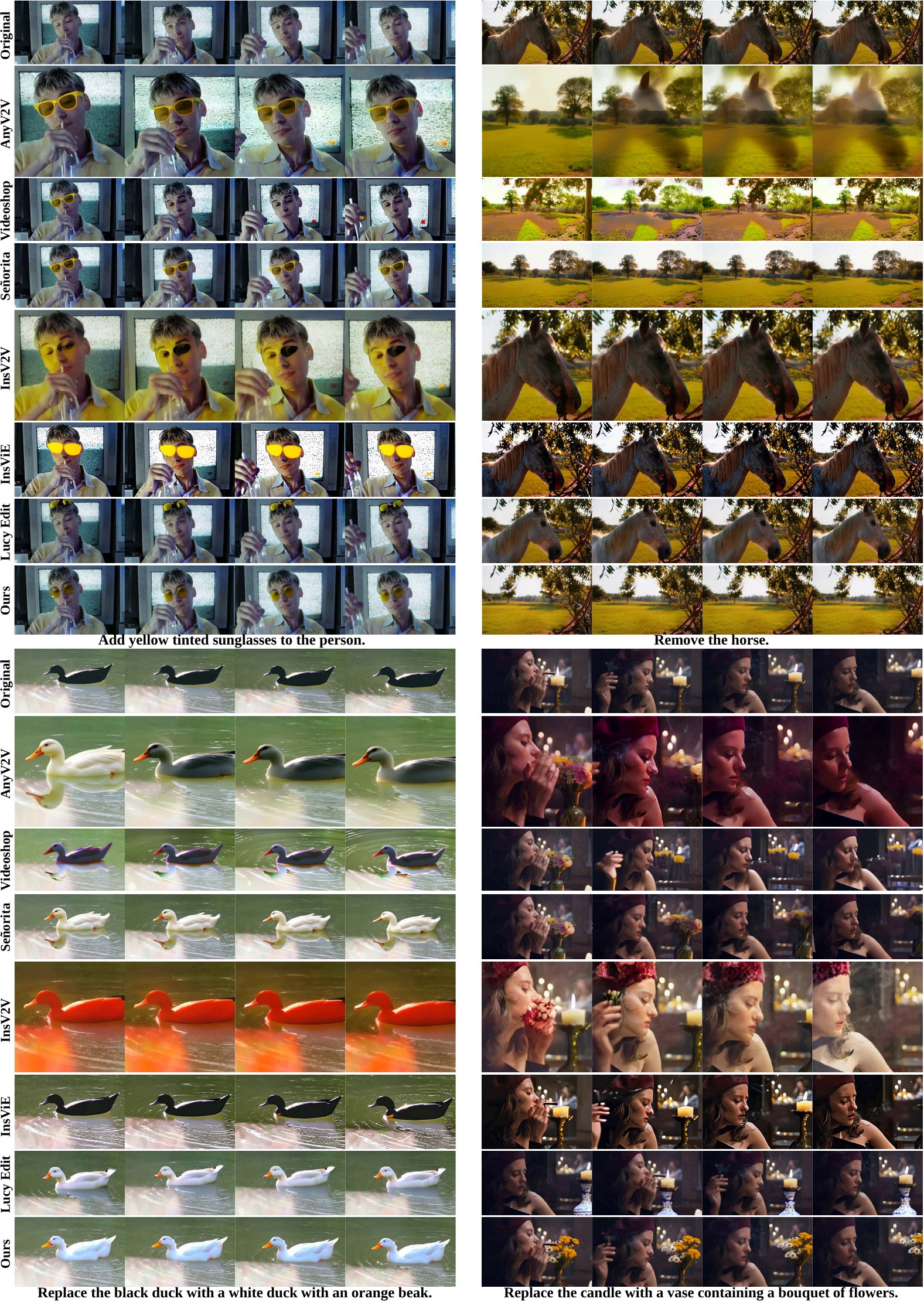}
\vspace{-0.8em}
\caption{Comparison results with the instruction-based methods. For methods that require the first edited frame, the first frame is obtained using Qwen-Image-Edit \citep{wu2025qwen}.}
\vspace{-0.8em}
\label{fig:comparison}
\end{figure}

\begin{table}[h]
\vspace{-1.5em}
\caption{Comparison results with instruction-based video editing methods. 
Best results in each column are highlighted in \textbf{bold}, and the second best are \underline{underlined}. The Overall score represents the average of I\_F, O\_P, and E\_Q.}
\label{tab:comparison}
\centering
\resizebox{1.0\linewidth}{!}
{
\begin{tabular}{lcccccccccccc}
    \toprule
    \multirow{2}{*}{Method} & \multirow{2}{*}{Extra Model}
      & \multicolumn{6}{c}{Editing Instruction}
      & \multicolumn{5}{c}{Video Quality} \\
    \cmidrule(lr){3-9} \cmidrule(lr){10-13}
      & & CLIP & Pick & I\_F & O\_P
      & E\_Q & S\_R & Overall
      & S\_C & B\_C
      & M\_S & T\_F \\
    \midrule
    AnyV2V \citep{ku2024anyv2v} & \multirow{3}{*}{+InsP2P}
      & 0.2534 & 19.6328 & 2.5896 & 3.2428 & 3.3526 & 25.43\%
      & 3.0617 & 0.9405 & 0.9509 & 0.9826 & 0.9750 \\
    Videoshop \citep{fan2024videoshop} &
      & 0.2425 & 19.3602 & 2.5260 & 2.5549 & 2.5549 & 23.70\%
      & 2.5453 & 0.9449 & 0.9535 & 0.9742 & 0.9550 \\
    Senorita-2M \citep{zi2025se} &
      & 0.2443 & 19.5202 & 2.5954 & 2.9191 & 3.0058 & 23.12\%
      & 2.8401 & 0.9480 & 0.9616 & 0.9910 & 0.9813 \\
    \cmidrule(lr){1-13}
    AnyV2V \citep{ku2024anyv2v} & \multirow{3}{*}{+Qwen}
      & 0.2678 & 20.0900 & 3.8439 & 3.9422 & 3.3526 & \underline{64.16\%}
      & 3.7129 & 0.9088 & 0.9292 & 0.9795 & 0.9719 \\
    Videoshop \citep{fan2024videoshop} &
      & 0.2605 & 19.7915 & 3.2890 & 3.4046 & 2.9075 & 37.57\%
      & 3.2004 & 0.9273 & 0.9419 & 0.9752 & 0.9571 \\
    Senorita-2M \citep{zi2025se} &
      & \textbf{0.2729} & \textbf{20.4642} & \underline{4.0549} & \underline{4.0809} & 3.5260 & \textbf{78.03\%}
      & \underline{3.8873} & 0.9601 & \underline{0.9651} & 0.9919 & 0.9827 \\
    \midrule
    InsV2V \citep{cheng2023consistent} & \multirow{4}{*}{-}
      & 0.2658 & 19.7429 & 2.3988 & 4.0694 & 3.4046 & 30.06\%
      & 3.2909 & 0.9630 & \textbf{0.9708} & 0.9873 & 0.9779 \\
    InsViE-1M \citep{wu2025insvie} &
      & 0.2489 & 19.5580 & 1.8902 & 3.2312 & 2.5376 & 18.50\%
      & 2.5530 & \underline{0.9675} & 0.9565 & 0.9810 & 0.9572 \\
    Lucy Edit Dev \citep{Lucy_Edit} &
      & 0.2531 & 19.8444 & 2.2543 & \underline{4.1387} & \underline{3.6474} & 24.28\%
      & 3.3468 & 0.9668 & 0.9459 & \underline{0.9922} & \underline{0.9838} \\
    Ours &
      & \underline{0.2701} & \underline{20.2516} & \textbf{4.5491} & \textbf{4.3064} & \textbf{4.2081} & \textbf{78.03\%}
      & \textbf{4.3545} & \textbf{0.9795} & 0.9636 & \textbf{0.9941} & \textbf{0.9867} \\
    \bottomrule
\end{tabular}
}
\vspace{-0.5em}
\end{table}

As shown in Table \ref{tab:comparison}, our method achieves the best performance on most metrics. For instruction-related metrics, it obtains the highest scores in Instruction Following, Original Video Preservation, Editing Quality, Success Ratio, and Overall Score, while ranking second in the CLIP and Pick scores. Our model achieves a 12\% improvement in editing instruction following and a 15\% improvement in editing quality. Although Señorita-2M+Qwen-Image-Edit \citep{zi2025se,wu2025qwen} achieves the same Success Ratio, its Editing Quality is lower than ours. This suggests that it fails to effectively propagate prior information from the first edited frame to subsequent frames. Consequently, both its Instruction Following and Editing Quality remain inferior to ours. Lucy Edit Dev \citep{Lucy_Edit}, which is based on Wan Video \citep{wan2025wan}, achieves good performance on Original Video Preservation and Editing Quality, but underperforms on Success Ratio.  

Regarding video generation quality, our method achieves the best results in Subject Consistency, Motion Smoothness, and Temporal Flickering, benefiting from the strong foundation model and the high-quality curated data used for training. The Background Consistency metric is also competitive, ranking third among all methods. Qualitative comparisons in Figure \ref{fig:comparison} further demonstrate that our method produces more precise and natural editing results than other instruction-based approaches.

\subsection{Ablation}
\textbf{Ablation on training strategies.} In the ablation study, we evaluate the effectiveness of the proposed strategy of first pretraining on video clip data and then performing SFT with a small amount of editing data. Three models are trained under different settings:  
(1) the model trained with the proposed strategy;  
(2) the model pretrained on (approximately 3M) editing data and then fine-tuned on the same (about 150K) SFT data, representing the standard training paradigm of editing models;
(3) the model trained directly on the same editing data for SFT without a pretraining stage, serving to examine whether a small amount of editing data alone is sufficient to train a high-quality model. Instruction Following, Original Video Preservation, Editing Quality, and Success Ratio are reported.
\begin{table}[h]
\vspace{-0.5em}
\caption{Ablation study on different training strategies.}
\label{tab:ablation}
\centering
\resizebox{0.70\linewidth}{!}
{
\begin{tabular}{lcccc}
    \toprule
    Setting & I\_F & O\_P & E\_Q & S\_R \\
    \midrule
    Pretraining Clip Data + SFT Edit Data (Ours) & 4.5491 & 4.3064 & 4.2081 & 78.03\% \\
    Pretraining Edit Data + SFT Edit Data & 4.3815 & 4.0809 & 4.0520 & 72.25\% \\
    w/o Pretraining + SFT Edit Data & 3.9191 & 3.6647 & 3.7746 & 56.65\% \\
    \bottomrule
\end{tabular}
}
\vspace{-0.5em}
\end{table}

As shown in Table \ref{tab:ablation}, the standard training paradigm, which relies entirely on editing data for both pretraining and SFT, fails to produce a higher-quality model than our proposed strategy. This limitation arises because training solely on synthetic data inevitably introduces artifacts, while the instruction labels of synthetic datasets are often inaccurate, degrading the final performance. In addition, this paradigm requires a very large volume of editing data (about 3M). The generation and filtering of such large-scale synthetic datasets demand substantial computational resources. The results also confirm that training with only a small amount of SFT editing data (about 150K) is insufficient to obtain satisfactory performance, as the model overfits and collapses. These findings highlight the necessity of the pretraining stage and the effectiveness of the proposed two-stage training strategy.  

\begin{figure}[b]
\centering
\includegraphics[width=0.99\linewidth]{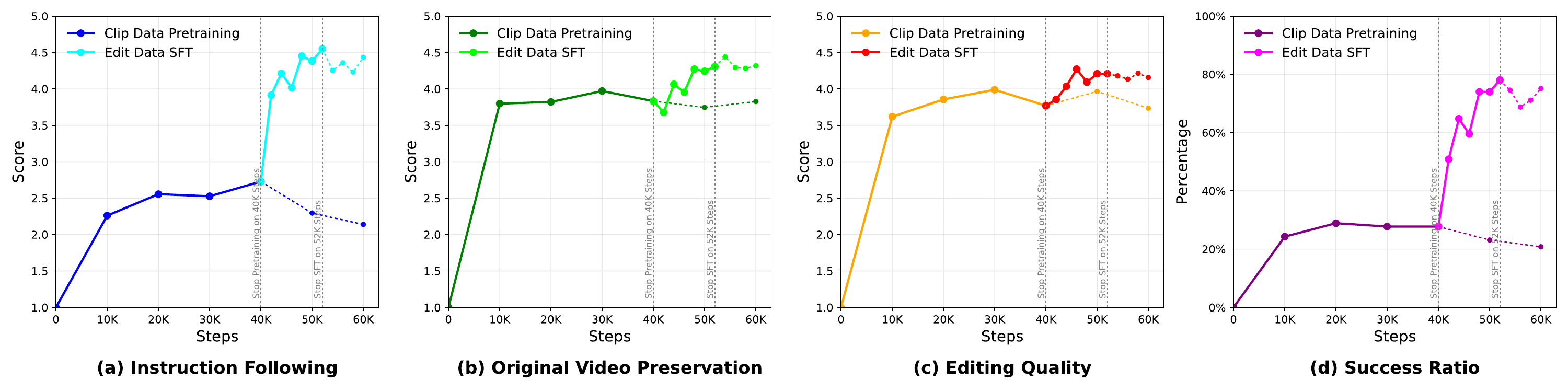}
\vspace{-0.8em}
\caption{Model performance with varying training steps and data. Dashed lines indicate evaluation results after continued training. The darker line in each sub-figure shows the pretrained model, which acquires basic editing capabilities from clip data, while the darker dashed line after 40K steps illustrates that prolonged pretraining causes model degradation. The lighter line demonstrates that the model improves rapidly with only a few SFT steps, and the lighter dashed line indicates that about 12K SFT steps are sufficient to fine-tune a high-quality editing model.}
\vspace{-0.8em}
\label{fig:ab}
\end{figure}
\begin{figure}[h]
\centering
\vspace{-0.8em}
\includegraphics[width=0.95\linewidth]{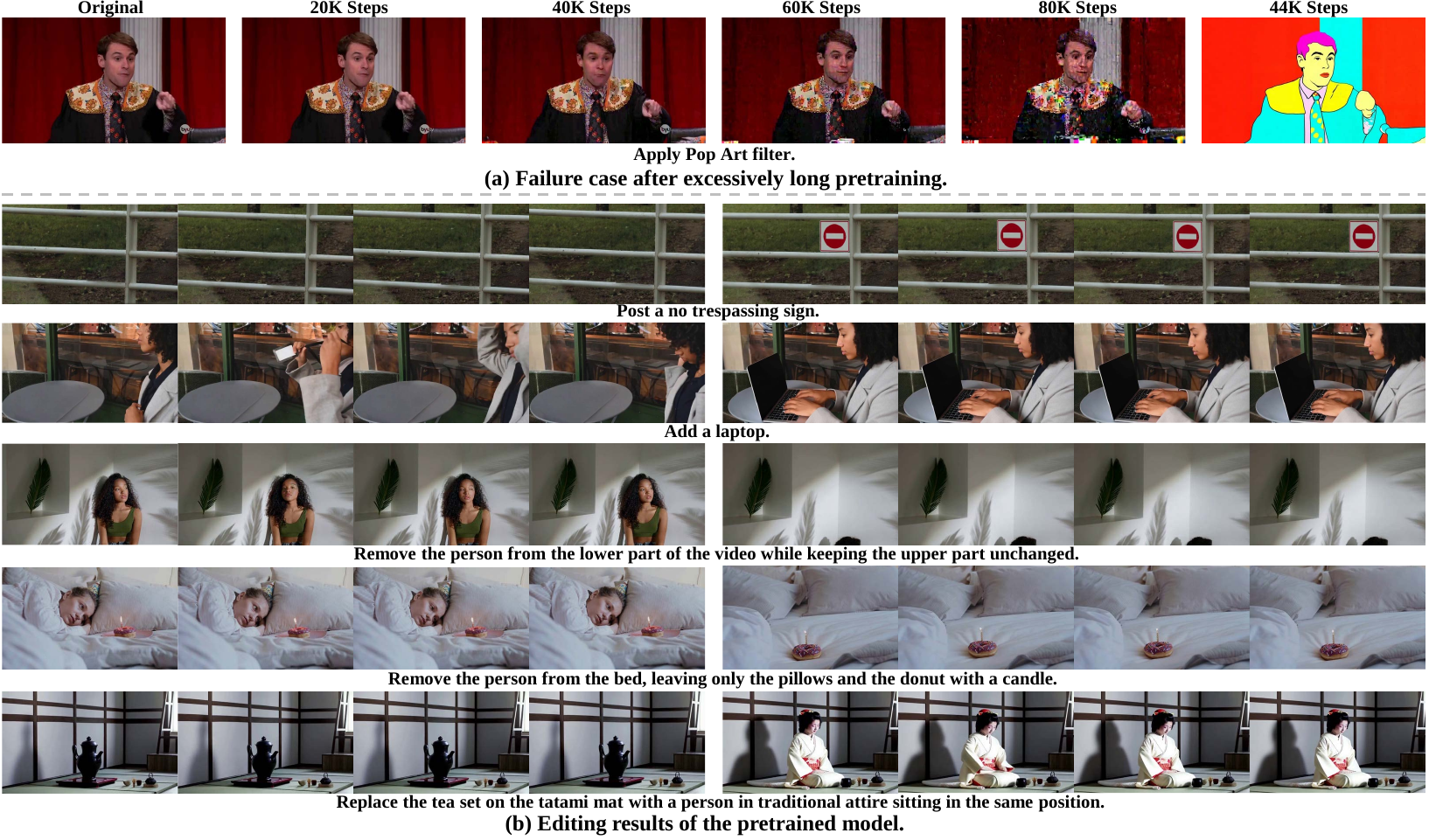}
\vspace{-1.0em}
\caption{Visualization results of the pretrained model. (a) shows that excessive pretraining leads to overfitting and collapse on unseen editing types. However, with a small number of SFT steps on editing data, the pretrained model quickly learns new editing types. (b) demonstrates that the pretrained model acquires basic editing operations, such as addition, removal, and replacement, using only video clip data. The left are the original videos, and the right are the edited results.}
\vspace{-1.5em}
\label{fig:pretrain}
\end{figure}

\textbf{Ablation on Training Steps and Data.}  
In this part, we investigate the impact of different amounts of clip data and SFT data on training. In the pretraining stage, the batch size is set to 32. As shown in Figure \ref{fig:ab}, after 40K training steps (corresponding to 1.28M clips, evaluated every 10K steps), the model learns to preserve original video content and maintain good video quality, but it cannot yet respond precisely to editing instructions. Some edited results generated by the pretrained model are shown in Figure \ref{fig:pretrain}(b). These results indicate that the model has already acquired some basic editing concepts from the video clips. However, due to the absence of editing data during training, the pretrained model cannot generate results that strictly follow the timeline of the original videos.  

We also observe limitations caused by excessively long pretraining. Figure \ref{fig:pretrain}(a) presents the same test case edited by models pretrained with different numbers of steps. Notably, this case involves style editing, which cannot be learned from clip data. The results in Figure \ref{fig:pretrain}(a) show that with prolonged pretraining (after 60K steps), the model tends to overfit to the clip data and eventually collapse. The dashed lines of pretraining in Figure \ref{fig:ab} also validate this degradation phenomenon. Therefore, we terminate pretraining at 40K steps.  

The impact of the amount of SFT data is also examined. In this stage, the batch size is set to 12. As shown in Figure \ref{fig:ab}, the model learns quickly from the SFT data. After only 4K training steps (48K editing pairs, evaluated every 2K steps), the model is able to perform accurate video editing. Furthermore, we observe that 120K–144K editing pairs are sufficient to train a high-quality model, indicating that only a relatively small amount of editing data is required in the SFT stage. Figure \ref{fig:pretrain}(a) also shows the style editing results at 44K steps. Even though the clip data does not contain any style editing examples, the model can acquire new editing capabilities within a few SFT steps.

\section{Conclusion}
In this paper, we present a data-efficient training strategy for instruction-based video editing. 
Our method combines pretraining on video clips with SFT on a relatively small amount of editing data. The pretraining stage generalizes the model with basic editing capabilities, and the SFT stage extends the editing type and improves video quality.
We introduce a data pipeline for curating clip data and synthesizing high-quality editing data, and further adapt an in-context instruction-based video editing model. 
Extensive experiments demonstrate that the proposed approach significantly reduces reliance on large-scale synthetic editing datasets, while achieving superior performance in both editing instruction following and video generation quality compared with existing approaches.

\bibliography{iclr2026_conference}

\begin{thebibliography}{51}
\providecommand{\natexlab}[1]{#1}
\providecommand{\url}[1]{\texttt{#1}}
\expandafter\ifx\csname urlstyle\endcsname\relax
  \providecommand{\doi}[1]{doi: #1}\else
  \providecommand{\doi}{doi: \begingroup \urlstyle{rm}\Url}\fi

\bibitem[Bai et~al.(2025)Bai, Chen, Liu, Wang, Ge, Song, Dang, Wang, Wang, Tang, et~al.]{bai2025qwen2}
Shuai Bai, Keqin Chen, Xuejing Liu, Jialin Wang, Wenbin Ge, Sibo Song, Kai Dang, Peng Wang, Shijie Wang, Jun Tang, et~al.
\newblock Qwen2. 5-vl technical report.
\newblock \emph{arXiv preprint arXiv:2502.13923}, 2025.

\bibitem[Brooks et~al.(2023)Brooks, Holynski, and Efros]{brooks2023instructpix2pix}
Tim Brooks, Aleksander Holynski, and Alexei~A Efros.
\newblock Instructpix2pix: Learning to follow image editing instructions.
\newblock In \emph{Proceedings of the IEEE/CVF conference on computer vision and pattern recognition}, pp.\  18392--18402, 2023.

\bibitem[Chen et~al.(2016)Chen, Xu, Zhang, and Guestrin]{chen2016training}
Tianqi Chen, Bing Xu, Chiyuan Zhang, and Carlos Guestrin.
\newblock Training deep nets with sublinear memory cost.
\newblock \emph{arXiv preprint arXiv:1604.06174}, 2016.

\bibitem[Cheng et~al.(2023)Cheng, Xiao, and He]{cheng2023consistent}
Jiaxin Cheng, Tianjun Xiao, and Tong He.
\newblock Consistent video-to-video transfer using synthetic dataset.
\newblock \emph{arXiv preprint arXiv:2311.00213}, 2023.

\bibitem[Cong et~al.(2023)Cong, Xu, Simon, Chen, Ren, Xie, Perez-Rua, Rosenhahn, Xiang, and He]{cong2023flatten}
Yuren Cong, Mengmeng Xu, Christian Simon, Shoufa Chen, Jiawei Ren, Yanping Xie, Juan-Manuel Perez-Rua, Bodo Rosenhahn, Tao Xiang, and Sen He.
\newblock Flatten: optical flow-guided attention for consistent text-to-video editing.
\newblock \emph{arXiv preprint arXiv:2310.05922}, 2023.

\bibitem[DecartAI(2025)]{Lucy_Edit}
DecartAI.
\newblock Lucy edit: Open-weight text-guided video editing.
\newblock \url{https://huggingface.co/decart-ai/Lucy-Edit-Dev}, 2025.

\bibitem[Fan et~al.(2024)Fan, Bhattad, and Krishna]{fan2024videoshop}
Xiang Fan, Anand Bhattad, and Ranjay Krishna.
\newblock Videoshop: Localized semantic video editing with noise-extrapolated diffusion inversion.
\newblock In \emph{European Conference on Computer Vision}, pp.\  232--250. Springer, 2024.

\bibitem[Gao et~al.(2025{\natexlab{a}})Gao, Ding, Cai, Huang, Wang, and Xue]{gao2025lora}
Chenjian Gao, Lihe Ding, Xin Cai, Zhanpeng Huang, Zibin Wang, and Tianfan Xue.
\newblock Lora-edit: Controllable first-frame-guided video editing via mask-aware lora fine-tuning.
\newblock \emph{arXiv preprint arXiv:2506.10082}, 2025{\natexlab{a}}.

\bibitem[Gao et~al.(2025{\natexlab{b}})Gao, Guo, Hoang, Huang, Jiang, Kong, Li, Li, Li, Li, et~al.]{gao2025seedance}
Yu~Gao, Haoyuan Guo, Tuyen Hoang, Weilin Huang, Lu~Jiang, Fangyuan Kong, Huixia Li, Jiashi Li, Liang Li, Xiaojie Li, et~al.
\newblock Seedance 1.0: Exploring the boundaries of video generation models.
\newblock \emph{arXiv preprint arXiv:2506.09113}, 2025{\natexlab{b}}.

\bibitem[Geyer et~al.(2023)Geyer, Bar-Tal, Bagon, and Dekel]{geyer2023tokenflow}
Michal Geyer, Omer Bar-Tal, Shai Bagon, and Tali Dekel.
\newblock Tokenflow: Consistent diffusion features for consistent video editing.
\newblock \emph{arXiv preprint arXiv:2307.10373}, 2023.

\bibitem[Google(2025)]{Gemini-2.5}
Google.
\newblock Gemini-2.5.
\newblock \url{https://deepmind.google/models/gemini/}, 2025.

\bibitem[Hu et~al.(2024)Hu, Zhong, Wang, Jiang, Tian, Yang, Wan, and Zhang]{hu2024vivid}
Jiahao Hu, Tianxiong Zhong, Xuebo Wang, Boyuan Jiang, Xingye Tian, Fei Yang, Pengfei Wan, and Di~Zhang.
\newblock Vivid-10m: A dataset and baseline for versatile and interactive video local editing.
\newblock \emph{arXiv preprint arXiv:2411.15260}, 2024.

\bibitem[Huang et~al.(2024)Huang, He, Yu, Zhang, Si, Jiang, Zhang, Wu, Jin, Chanpaisit, et~al.]{huang2024vbench}
Ziqi Huang, Yinan He, Jiashuo Yu, Fan Zhang, Chenyang Si, Yuming Jiang, Yuanhan Zhang, Tianxing Wu, Qingyang Jin, Nattapol Chanpaisit, et~al.
\newblock Vbench: Comprehensive benchmark suite for video generative models.
\newblock In \emph{Proceedings of the IEEE/CVF Conference on Computer Vision and Pattern Recognition}, pp.\  21807--21818, 2024.

\bibitem[HunyuanLoom(2025)]{HunyuanLoom}
HunyuanLoom.
\newblock Hunyuanloom.
\newblock \url{https://github.com/logtd/ComfyUI-HunyuanLoom}, 2025.

\bibitem[Jacobs et~al.(2023)Jacobs, Tanaka, Zhang, Zhang, Song, Rajbhandari, and He]{jacobs2023deepspeed}
Sam~Ade Jacobs, Masahiro Tanaka, Chengming Zhang, Minjia Zhang, Shuaiwen~Leon Song, Samyam Rajbhandari, and Yuxiong He.
\newblock Deepspeed ulysses: System optimizations for enabling training of extreme long sequence transformer models.
\newblock \emph{arXiv preprint arXiv:2309.14509}, 2023.

\bibitem[Jiang et~al.(2025)Jiang, Han, Mao, Zhang, Pan, and Liu]{jiang2025vace}
Zeyinzi Jiang, Zhen Han, Chaojie Mao, Jingfeng Zhang, Yulin Pan, and Yu~Liu.
\newblock Vace: All-in-one video creation and editing.
\newblock \emph{arXiv preprint arXiv:2503.07598}, 2025.

\bibitem[Kara et~al.(2024)Kara, Kurtkaya, Yesiltepe, Rehg, and Yanardag]{kara2024rave}
Ozgur Kara, Bariscan Kurtkaya, Hidir Yesiltepe, James~M Rehg, and Pinar Yanardag.
\newblock Rave: Randomized noise shuffling for fast and consistent video editing with diffusion models.
\newblock In \emph{Proceedings of the IEEE/CVF Conference on Computer Vision and Pattern Recognition}, pp.\  6507--6516, 2024.

\bibitem[Kirstain et~al.(2023)Kirstain, Polyak, Singer, Matiana, Penna, and Levy]{kirstain2023pick}
Yuval Kirstain, Adam Polyak, Uriel Singer, Shahbuland Matiana, Joe Penna, and Omer Levy.
\newblock Pick-a-pic: An open dataset of user preferences for text-to-image generation.
\newblock \emph{Advances in neural information processing systems}, 36:\penalty0 36652--36663, 2023.

\bibitem[Kong et~al.(2024)Kong, Tian, Zhang, Min, Dai, Zhou, Xiong, Li, Wu, Zhang, et~al.]{kong2024hunyuanvideo}
Weijie Kong, Qi~Tian, Zijian Zhang, Rox Min, Zuozhuo Dai, Jin Zhou, Jiangfeng Xiong, Xin Li, Bo~Wu, Jianwei Zhang, et~al.
\newblock Hunyuanvideo: A systematic framework for large video generative models.
\newblock \emph{arXiv preprint arXiv:2412.03603}, 2024.

\bibitem[Korthikanti et~al.(2023)Korthikanti, Casper, Lym, McAfee, Andersch, Shoeybi, and Catanzaro]{korthikanti2023reducing}
Vijay~Anand Korthikanti, Jared Casper, Sangkug Lym, Lawrence McAfee, Michael Andersch, Mohammad Shoeybi, and Bryan Catanzaro.
\newblock Reducing activation recomputation in large transformer models.
\newblock \emph{Proceedings of Machine Learning and Systems}, 5:\penalty0 341--353, 2023.

\bibitem[Ku et~al.(2024)Ku, Wei, Ren, Yang, and Chen]{ku2024anyv2v}
Max Ku, Cong Wei, Weiming Ren, Harry Yang, and Wenhu Chen.
\newblock Anyv2v: A tuning-free framework for any video-to-video editing tasks.
\newblock \emph{arXiv preprint arXiv:2403.14468}, 2024.

\bibitem[Kulikov et~al.(2024)Kulikov, Kleiner, Huberman-Spiegelglas, and Michaeli]{kulikov2024flowedit}
Vladimir Kulikov, Matan Kleiner, Inbar Huberman-Spiegelglas, and Tomer Michaeli.
\newblock Flowedit: Inversion-free text-based editing using pre-trained flow models.
\newblock \emph{arXiv preprint arXiv:2412.08629}, 2024.

\bibitem[Labs et~al.(2025)Labs, Batifol, Blattmann, Boesel, Consul, Diagne, Dockhorn, English, English, Esser, et~al.]{labs2025flux}
Black~Forest Labs, Stephen Batifol, Andreas Blattmann, Frederic Boesel, Saksham Consul, Cyril Diagne, Tim Dockhorn, Jack English, Zion English, Patrick Esser, et~al.
\newblock Flux. 1 kontext: Flow matching for in-context image generation and editing in latent space.
\newblock \emph{arXiv preprint arXiv:2506.15742}, 2025.

\bibitem[Li et~al.(2025)Li, Yang, Song, and Zhang]{li2025flowdirector}
Guangzhao Li, Yanming Yang, Chenxi Song, and Chi Zhang.
\newblock Flowdirector: Training-free flow steering for precise text-to-video editing.
\newblock \emph{arXiv preprint arXiv:2506.05046}, 2025.

\bibitem[Li et~al.(2021)Li, Xue, Baranwal, Li, and You]{li2021sequence}
Shenggui Li, Fuzhao Xue, Chaitanya Baranwal, Yongbin Li, and Yang You.
\newblock Sequence parallelism: Long sequence training from system perspective.
\newblock \emph{arXiv preprint arXiv:2105.13120}, 2021.

\bibitem[Li et~al.(2024)Li, Ma, Yang, and Yang]{li2024vidtome}
Xirui Li, Chao Ma, Xiaokang Yang, and Ming-Hsuan Yang.
\newblock Vidtome: Video token merging for zero-shot video editing.
\newblock In \emph{Proceedings of the IEEE/CVF Conference on Computer Vision and Pattern Recognition}, pp.\  7486--7495, 2024.

\bibitem[Lian et~al.(2024)Lian, Jacobs, Kurilenko, Tanaka, Bekman, Ruwase, and Zhang]{lian2024universal}
Xinyu Lian, Sam~Ade Jacobs, Lev Kurilenko, Masahiro Tanaka, Stas Bekman, Olatunji Ruwase, and Minjia Zhang.
\newblock Universal checkpointing: Efficient and flexible checkpointing for large scale distributed training.
\newblock \emph{arXiv preprint arXiv:2406.18820}, 2024.

\bibitem[Liu et~al.(2024)Liu, Zhang, Li, Lin, and Jia]{liu2024video}
Shaoteng Liu, Yuechen Zhang, Wenbo Li, Zhe Lin, and Jiaya Jia.
\newblock Video-p2p: Video editing with cross-attention control.
\newblock In \emph{Proceedings of the IEEE/CVF Conference on Computer Vision and Pattern Recognition}, pp.\  8599--8608, 2024.

\bibitem[Liu et~al.(2025)Liu, Han, Xing, Yin, Wang, Cheng, Liao, Wang, Fu, Han, et~al.]{liu2025step1x}
Shiyu Liu, Yucheng Han, Peng Xing, Fukun Yin, Rui Wang, Wei Cheng, Jiaqi Liao, Yingming Wang, Honghao Fu, Chunrui Han, et~al.
\newblock Step1x-edit: A practical framework for general image editing.
\newblock \emph{arXiv preprint arXiv:2504.17761}, 2025.

\bibitem[Ma et~al.(2025)Ma, Feng, Hu, Wang, Wang, Zheng, He, Zhu, Liu, He, et~al.]{ma2025controllable}
Yue Ma, Kunyu Feng, Zhongyuan Hu, Xinyu Wang, Yucheng Wang, Mingzhe Zheng, Xuanhua He, Chenyang Zhu, Hongyu Liu, Yingqing He, et~al.
\newblock Controllable video generation: A survey.
\newblock \emph{arXiv preprint arXiv:2507.16869}, 2025.

\bibitem[OpenAI(2025)]{GPT-5}
OpenAI.
\newblock Gpt-5.
\newblock \url{https://openai.com/gpt-5/}, 2025.

\bibitem[Pexels(2025)]{Pexels}
Pexels.
\newblock Pexels.
\newblock \url{https://www.pexels.com/}, 2025.

\bibitem[Pont-Tuset et~al.(2017)Pont-Tuset, Perazzi, Caelles, Arbel{\'a}ez, Sorkine-Hornung, and Van~Gool]{pont20172017}
Jordi Pont-Tuset, Federico Perazzi, Sergi Caelles, Pablo Arbel{\'a}ez, Alex Sorkine-Hornung, and Luc Van~Gool.
\newblock The 2017 davis challenge on video object segmentation.
\newblock \emph{arXiv preprint arXiv:1704.00675}, 2017.

\bibitem[Qi et~al.(2023)Qi, Cun, Zhang, Lei, Wang, Shan, and Chen]{qi2023fatezero}
Chenyang Qi, Xiaodong Cun, Yong Zhang, Chenyang Lei, Xintao Wang, Ying Shan, and Qifeng Chen.
\newblock Fatezero: Fusing attentions for zero-shot text-based video editing.
\newblock In \emph{Proceedings of the IEEE/CVF International Conference on Computer Vision}, pp.\  15932--15942, 2023.

\bibitem[Qin et~al.(2024)Qin, Li, Tang, Chua, and Zhuang]{qin2024instructvid2vid}
Bosheng Qin, Juncheng Li, Siliang Tang, Tat-Seng Chua, and Yueting Zhuang.
\newblock Instructvid2vid: Controllable video editing with natural language instructions.
\newblock In \emph{2024 IEEE International Conference on Multimedia and Expo (ICME)}, pp.\  1--6. IEEE, 2024.

\bibitem[Radford et~al.(2021)Radford, Kim, Hallacy, Ramesh, Goh, Agarwal, Sastry, Askell, Mishkin, Clark, et~al.]{radford2021learning}
Alec Radford, Jong~Wook Kim, Chris Hallacy, Aditya Ramesh, Gabriel Goh, Sandhini Agarwal, Girish Sastry, Amanda Askell, Pamela Mishkin, Jack Clark, et~al.
\newblock Learning transferable visual models from natural language supervision.
\newblock In \emph{International conference on machine learning}, pp.\  8748--8763. PmLR, 2021.

\bibitem[Ravi et~al.(2024)Ravi, Gabeur, Hu, Hu, Ryali, Ma, Khedr, R{\"a}dle, Rolland, Gustafson, et~al.]{ravi2024sam}
Nikhila Ravi, Valentin Gabeur, Yuan-Ting Hu, Ronghang Hu, Chaitanya Ryali, Tengyu Ma, Haitham Khedr, Roman R{\"a}dle, Chloe Rolland, Laura Gustafson, et~al.
\newblock Sam 2: Segment anything in images and videos.
\newblock \emph{arXiv preprint arXiv:2408.00714}, 2024.

\bibitem[Ren et~al.(2024)Ren, Liu, Zeng, Lin, Li, Cao, Chen, Huang, Chen, Yan, et~al.]{ren2024grounded}
Tianhe Ren, Shilong Liu, Ailing Zeng, Jing Lin, Kunchang Li, He~Cao, Jiayu Chen, Xinyu Huang, Yukang Chen, Feng Yan, et~al.
\newblock Grounded sam: Assembling open-world models for diverse visual tasks.
\newblock \emph{arXiv preprint arXiv:2401.14159}, 2024.

\bibitem[SceneDetect(2025)]{SceneDetect}
SceneDetect.
\newblock Scenedetect.
\newblock \url{https://github.com/Breakthrough/PySceneDetect}, 2025.

\bibitem[Wan et~al.(2025)Wan, Wang, Ai, Wen, Mao, Xie, Chen, Yu, Zhao, Yang, et~al.]{wan2025wan}
Team Wan, Ang Wang, Baole Ai, Bin Wen, Chaojie Mao, Chen-Wei Xie, Di~Chen, Feiwu Yu, Haiming Zhao, Jianxiao Yang, et~al.
\newblock Wan: Open and advanced large-scale video generative models.
\newblock \emph{arXiv preprint arXiv:2503.20314}, 2025.

\bibitem[Wang et~al.(2025{\natexlab{a}})Wang, Wang, Wan, Huang, Hu, Jia, Nie, Li, Chen, Chen, et~al.]{wang2025step}
Bin Wang, Bojun Wang, Changyi Wan, Guanzhe Huang, Hanpeng Hu, Haonan Jia, Hao Nie, Mingliang Li, Nuo Chen, Siyu Chen, et~al.
\newblock Step-3 is large yet affordable: Model-system co-design for cost-effective decoding.
\newblock \emph{arXiv preprint arXiv:2507.19427}, 2025{\natexlab{a}}.

\bibitem[Wang et~al.(2025{\natexlab{b}})Wang, Fan, Liu, Song, Wang, and Xu]{wang2025consistent}
Ge~Wang, Songlin Fan, Hangxu Liu, Quanjian Song, Hewei Wang, and Jinfeng Xu.
\newblock Consistent video editing as flow-driven image-to-video generation.
\newblock \emph{arXiv preprint arXiv:2506.07713}, 2025{\natexlab{b}}.

\bibitem[Wang et~al.(2025{\natexlab{c}})Wang, Wang, Ma, Hu, Xu, and Guo]{wang2025videodirector}
Yukun Wang, Longguang Wang, Zhiyuan Ma, Qibin Hu, Kai Xu, and Yulan Guo.
\newblock Videodirector: Precise video editing via text-to-video models.
\newblock In \emph{Proceedings of the Computer Vision and Pattern Recognition Conference}, pp.\  2589--2598, 2025{\natexlab{c}}.

\bibitem[Wu et~al.(2025{\natexlab{a}})Wu, Li, Zhou, Lin, Gao, Yan, Yin, Bai, Xu, Chen, et~al.]{wu2025qwen}
Chenfei Wu, Jiahao Li, Jingren Zhou, Junyang Lin, Kaiyuan Gao, Kun Yan, Sheng-ming Yin, Shuai Bai, Xiao Xu, Yilei Chen, et~al.
\newblock Qwen-image technical report.
\newblock \emph{arXiv preprint arXiv:2508.02324}, 2025{\natexlab{a}}.

\bibitem[Wu et~al.(2023)Wu, Ge, Wang, Lei, Gu, Shi, Hsu, Shan, Qie, and Shou]{wu2023tune}
Jay~Zhangjie Wu, Yixiao Ge, Xintao Wang, Stan~Weixian Lei, Yuchao Gu, Yufei Shi, Wynne Hsu, Ying Shan, Xiaohu Qie, and Mike~Zheng Shou.
\newblock Tune-a-video: One-shot tuning of image diffusion models for text-to-video generation.
\newblock In \emph{Proceedings of the IEEE/CVF international conference on computer vision}, pp.\  7623--7633, 2023.

\bibitem[Wu et~al.(2025{\natexlab{b}})Wu, Chen, Li, Wang, Xie, and Zhang]{wu2025insvie}
Yuhui Wu, Liyi Chen, Ruibin Li, Shihao Wang, Chenxi Xie, and Lei Zhang.
\newblock Insvie-1m: Effective instruction-based video editing with elaborate dataset construction.
\newblock \emph{arXiv preprint arXiv:2503.20287}, 2025{\natexlab{b}}.

\bibitem[Xu et~al.(2018)Xu, Yang, Fan, Yue, Liang, Yang, and Huang]{xu2018youtube}
Ning Xu, Linjie Yang, Yuchen Fan, Dingcheng Yue, Yuchen Liang, Jianchao Yang, and Thomas Huang.
\newblock Youtube-vos: A large-scale video object segmentation benchmark.
\newblock \emph{arXiv preprint arXiv:1809.03327}, 2018.

\bibitem[Zhang et~al.(2025)Zhang, Feng, Yan, Zhang, Zhang, Zhong, Zhang, and Ma]{zhang2025instructvedit}
Chi Zhang, Chengjian Feng, Feng Yan, Qiming Zhang, Mingjin Zhang, Yujie Zhong, Jing Zhang, and Lin Ma.
\newblock Instructvedit: A holistic approach for instructional video editing.
\newblock \emph{arXiv preprint arXiv:2503.17641}, 2025.

\bibitem[Zhang et~al.(2024)Zhang, Dai, Qin, and Wang]{zhang2024effived}
Zhenghao Zhang, Zuozhuo Dai, Long Qin, and Weizhi Wang.
\newblock Effived: Efficient video editing via text-instruction diffusion models.
\newblock \emph{arXiv preprint arXiv:2403.11568}, 2024.

\bibitem[Zhao et~al.(2023)Zhao, Gu, Varma, Luo, Huang, Xu, Wright, Shojanazeri, Ott, Shleifer, et~al.]{zhao2023pytorch}
Yanli Zhao, Andrew Gu, Rohan Varma, Liang Luo, Chien-Chin Huang, Min Xu, Less Wright, Hamid Shojanazeri, Myle Ott, Sam Shleifer, et~al.
\newblock Pytorch fsdp: experiences on scaling fully sharded data parallel.
\newblock \emph{arXiv preprint arXiv:2304.11277}, 2023.

\bibitem[Zi et~al.(2025)Zi, Ruan, Chen, Qi, Hao, Zhao, Huang, Liang, Xiao, and Wong]{zi2025se}
Bojia Zi, Penghui Ruan, Marco Chen, Xianbiao Qi, Shaozhe Hao, Shihao Zhao, Youze Huang, Bin Liang, Rong Xiao, and Kam-Fai Wong.
\newblock Se$\backslash$\~{} norita-2m: A high-quality instruction-based dataset for general video editing by video specialists.
\newblock \emph{arXiv preprint arXiv:2502.06734}, 2025.

\end{thebibliography}
\bibliographystyle{iclr2026_conference}

\appendix
\section*{Appendix}
\section{Training Details.}
\textbf{Pretraining Stage.}  
The model is initialized with the weights of HunyuanVideoT2V \citep{kong2024hunyuanvideo}. In the 240p pretraining stage, three resolution buckets are used: (240, 384, 77), (384, 240, 77), and (256, 256, 125), where the triplets denote height, width, and number of frames, respectively. The batch size is set to 32, and the 240p pretraining is conducted for approximately 30K iterations. In the subsequent 480p pretraining stage, the resolution buckets are set to (480, 768, 77), (768, 480, 77), and (512, 512, 125). The batch size remains 32, and training continues for an additional 10K iterations. In total, 40K iterations of pretraining are performed using only video clip data, with the learning rate fixed at $1\times10^{-5}$. This stage consumes about 1.28M video clips.  

\textbf{SFT Stage.}  
To accommodate the SFT editing data from multiple sources, seven resolution buckets are adopted: (480, 768, 77), (768, 480, 77), (576, 1024, 21), (1024, 576, 21), (336, 592, 29), (592, 336, 29), and (480, 480, 125). The global batch size is set to 12, and a mini-batch strategy is applied to balance the number of tokens across different buckets. SFT is conducted for 12K iterations on high-quality editing data, with the learning rate set to $1\times10^{-6}$. This stage consumes about 144K of editing data.  

\textbf{Overall Training.}  
The full training process consists of 52K iterations, with pretraining accounting for roughly 75\% of the total training time, and the remaining 25\% dedicated to SFT on a relatively small amount of high-quality editing data.  

\section{Infrastructure Optimization}
To enable scalable and efficient training, we adopt several system-level optimizations. First, advanced parallelism strategies are introduced to address the challenges posed by large-scale models with long contexts. Second, a fine-grained activation checkpointing scheme is applied to reduce memory usage and better balance computation with communication. Finally, distributed checkpointing is employed to ensure efficient and scalable saving and loading of training states. The details of these optimizations are described below.  

\textbf{Model Parallelism.}  
The large model size and extremely long sequence length necessitate multiple parallelism strategies for efficient training. We employ 3D parallelism, which scales along three dimensions: input sequences, data, and model parameters. For input sequences, Sequence Parallelism \citep{li2021sequence,korthikanti2023reducing,jacobs2023deepspeed} shards samples across sequence-parallel groups at the start of training. During attention computation, all-to-all communication distributes query, key, and value shards so that each worker processes the full sequence but only a subset of attention heads. After parallel computation, another all-to-all step aggregates the outputs, recombining both the attention heads and the sharded sequence dimension. For parameters, gradients, and optimizer states, we use Fully Sharded Data Parallelism \citep{zhao2023pytorch}, which applies full sharding within each shard group while replicating parameters across groups. This approach effectively implements data parallelism while reducing communication costs by limiting all-gather and reduce-scatter operations.  

\textbf{Activation Checkpointing.}  
Selective activation checkpointing \citep{chen2016training} is applied to minimize the number of layers requiring activation storage while maximizing GPU utilization, thereby improving training efficiency without excessive memory consumption.  

\textbf{Distributed Saving and Loading.}  
To support scalable training, we adopt a distributed checkpointing solution \citep{lian2024universal} that enables parallel saving and loading of partitioned states with high I/O efficiency. It also supports resharding across distributed checkpoints, providing flexibility to switch seamlessly between different training scales, numbers of ranks, and storage backends.  

\section{More Pretraining Results}
\begin{figure}[h]
\centering
\includegraphics[width=0.99\linewidth]{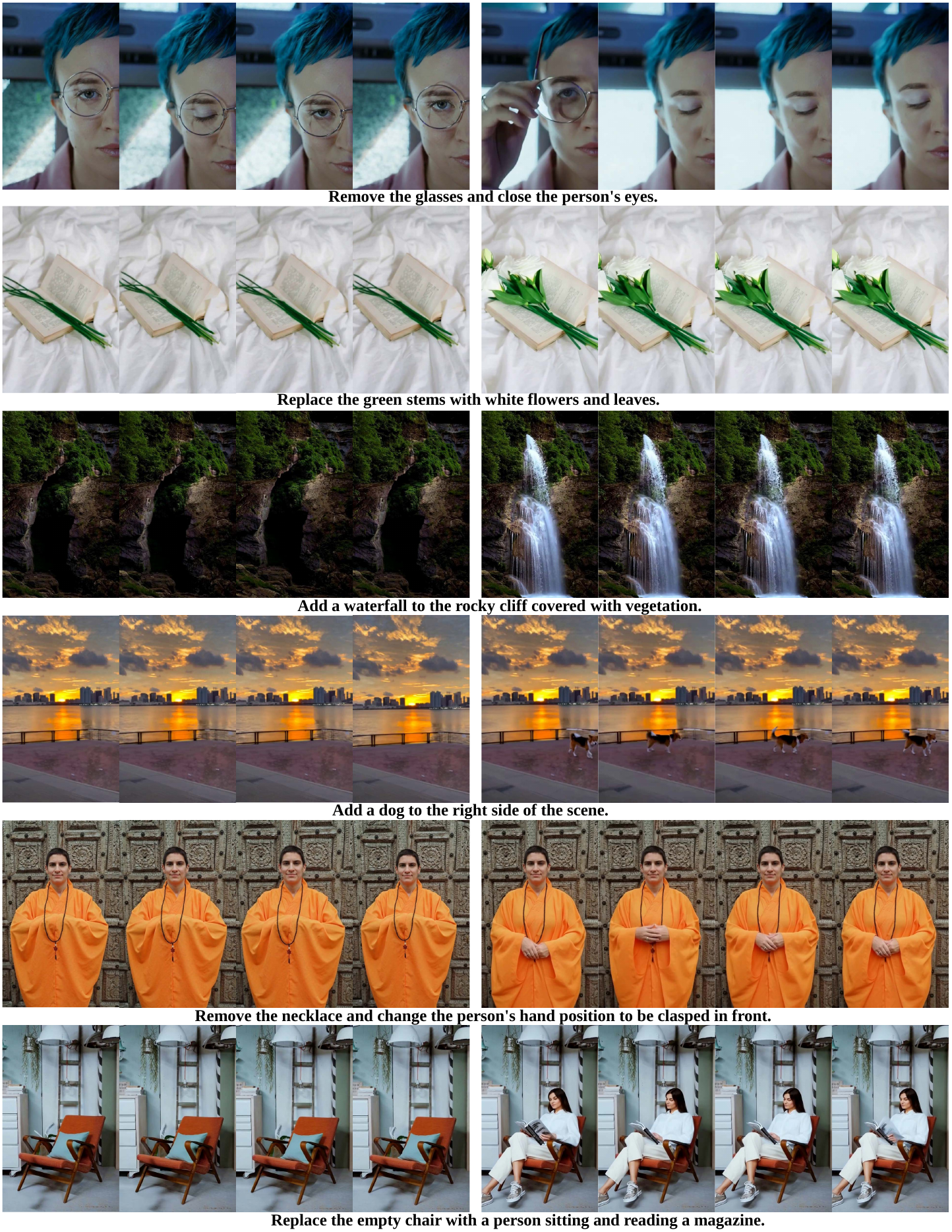}
\caption{Additional visualization results of the pretrained model. The results show that the model acquires basic editing operations using only video clip data pretraining. The left figures are the original video, and the right figures are the edited results by the pretrained model.}
\label{fig:pretrain_sup}
\end{figure}

Figure \ref{fig:pretrain_sup} presents several examples from the model pretrained solely on video clip data without any editing pairs. The results show that pretraining on clip data enables the model to acquire basic concepts of video editing. However, in the absence of explicit editing data, the model still produces noticeable errors, and the generated outputs do not strictly adhere to the given instructions. In our approach, pretraining is terminated at 40K steps, which provides a balance between learning basic editing concepts from clip data and avoiding model collapse.

\section{More Comparison Results}
\begin{figure}[h]
\centering
\includegraphics[width=0.99\linewidth]{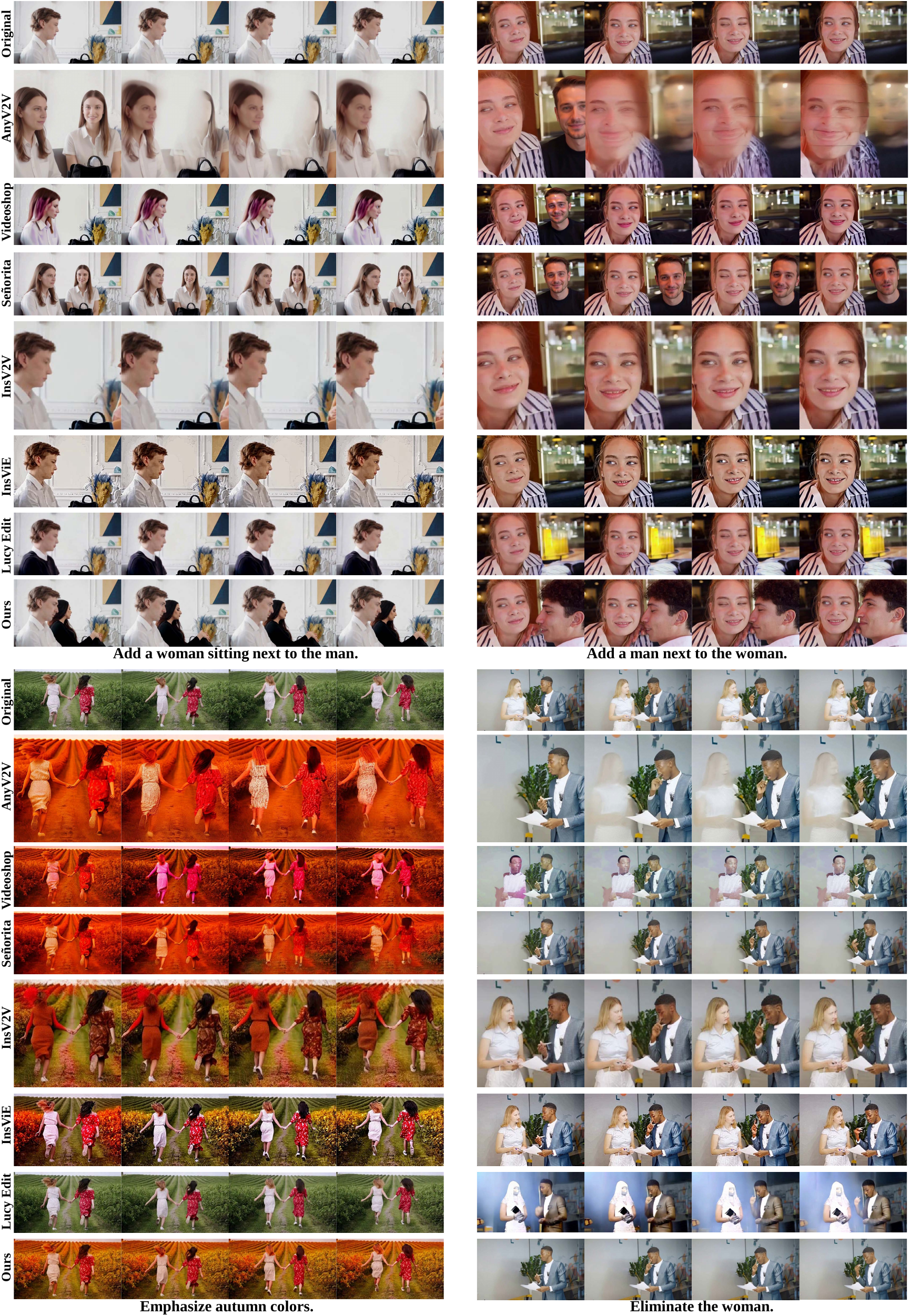}
\caption{More comparison results with instruction-based methods. For methods requiring the first edited frame, the frame is generated using Qwen-Image-Edit \citep{wu2025qwen}.}
\label{fig:comparison_1}
\end{figure}

\begin{figure}[h]
\centering
\includegraphics[width=0.99\linewidth]{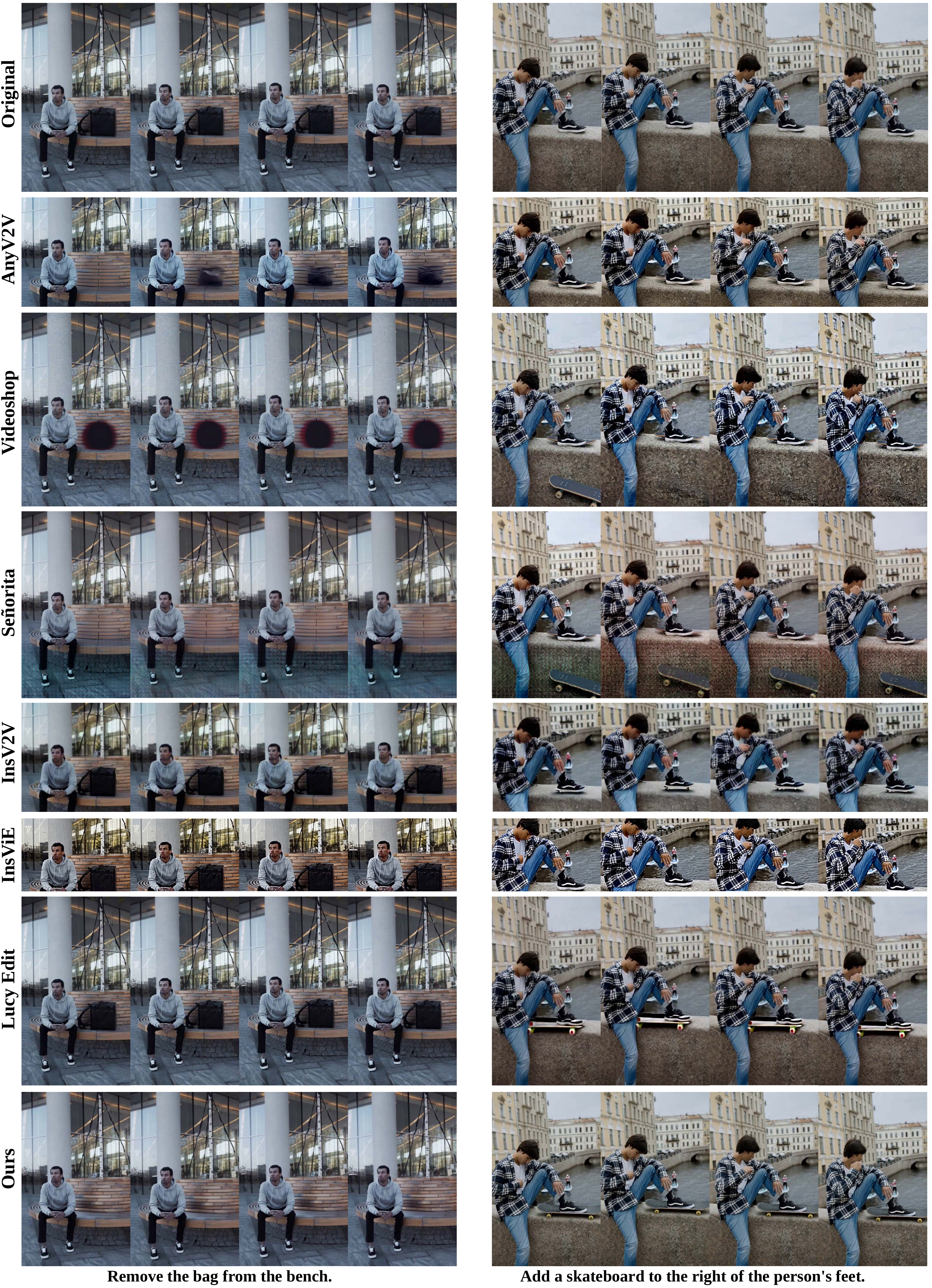}
\caption{More comparison results with instruction-based methods. For methods requiring the first edited frame, the frame is generated using Qwen-Image-Edit \citep{wu2025qwen}.}
\label{fig:comparison_3}
\end{figure}

In Figure \ref{fig:comparison_1}, \ref{fig:comparison_3}, more qualitative comparisons are shown. Our proposed video editing model achieves more accurate editing instruction following and generates high-quality video results.

\section{Editing Instruction Prompt}
During the pretraining stage, Step3 \citep{wang2025step} is employed as the captioning model. The Table \ref{tab:prompt_instruction_generation} shows the prompt used for generating editing instructions from video clip data.  

\section{Automatic Evaluation Prompts}
In the experiments, GPT-5 \citep{GPT-5} is used for the automated evaluation of editing metrics. Several representative prompts for guiding the large language model in performing automatic evaluation are presented in Tables \ref{tab:prompt_instruction_following}, \ref{tab:prompt_original_preservation}, \ref{tab:prompt_editing_quality}, \ref{tab:prompt_success_ratio}.

\section{Limitations and Future Work}
The in-context integration of original video tokens nearly doubles the sequence length, leading to quadratic computational costs in the full attention operation. However, much of the visual content within the original video tokens is redundant. In future work, we plan to explore effective approaches for compressing original video tokens and reducing the sequence length, thereby enabling more efficient instruction-based video editing.

\section{The Use of Large Language Models}
In this paper, large language models are primarily employed to correct grammatical errors and refine sentence structures, thereby improving the linguistic accuracy and clarity of expression, and helping the paper better conform to academic writing standards and enhance its overall readability.

\begin{table}[h]
\centering
\caption{Prompt template for video editing instruction generation.}
\label{tab:prompt_instruction_generation}
\renewcommand{\arraystretch}{1.3}
\begin{tabular}{|p{0.95\linewidth}|}
\hline
\textbf{Prompt Template: Video Editing Instruction Generation} \\
\hline
You are an advanced vision–language model tasked with generating precise video-editing instructions based on the original video and the edited video. \\[0.6em]

The following images are frames from the original video and the edited video: the left image shows the original frame, and the right image shows the edited frame. \\[0.6em]

You should generate a video-editing instruction describing the differences between the original and the edited version. \\[0.6em]

You must reply with only \textbf{one instruction representing the most significant editing operation, with no additional analysis text}. \\[0.6em]

The instruction should be concise and specific, focusing on the primary change made in the edited version. \\[0.6em]

Do not use words such as ``frame'', ``image'' or ``video'' in your response. \\
\hline
\end{tabular}
\end{table}

\begin{table}[h]
\centering
\caption{Prompt template for instruction following scoring.}
\label{tab:prompt_instruction_following}
\renewcommand{\arraystretch}{1.25}
\begin{tabular}{|p{0.95\linewidth}|}
\hline
\textbf{Prompt Template: Instruction Following Scoring} \\
\hline
You are an advanced vision–language model tasked with grading instruction adherence for a video edit. 
I will provide several paired frames from an original video (left) and its edited version (right), separated by a blank divider. \\[0.6em]

The editing instruction is: \verb|<instruct_prompt>| \\[0.6em]

Judge how well the edited video \textbf{follows the instruction only, without introducing unrelated edits}.
Score on a 1–5 scale and output only one digit: \\[0.6em]

\begin{tabular}{@{}ll}
5 & perfectly follows; all required changes are present; no unrelated changes \\
4 & mostly follows; minor missing details or minor unrelated changes \\
3 & partially follows; noticeable missing parts or some unrelated changes \\
2 & poorly follows; major missing parts or many unrelated changes \\
1 & does not follow; requested edits absent or mostly wrong \\
\end{tabular} \\[0.6em]

Output only one digit: 1, 2, 3, 4, or 5. No words, no punctuation, no explanation. \\
\hline
\end{tabular}
\end{table}

\begin{table}[h]
\centering
\caption{Prompt template for original video preservation scoring.}
\label{tab:prompt_original_preservation}
\renewcommand{\arraystretch}{1.25}
\begin{tabular}{|p{0.95\linewidth}|}
\hline
\textbf{Prompt Template: Original Video Preservation Scoring} \\
\hline
You are an advanced vision–language model acting as a video edit preservation rater. 
I will provide several sampled frame pairs from one original video (left) and its edited version (right), separated by a blank divider. \\[0.6em]

The editing instruction for this case is: \verb|<instruct_prompt>| \\[0.6em]

Your task:  
\begin{enumerate}
    \item From the instruction, infer which elements should be changed (the requested edits).
    \item For all other elements that are not requested to change (backgrounds, identities, layout, colors, etc.), 
          \textbf{judge how well they are preserved in the edited video compared with the original}.
\end{enumerate} \\[0.6em]

Rate only the preservation of the unedited parts on a 1–5 scale and output only one digit: \\[0.6em]

\begin{tabular}{@{}ll}
5 & unedited parts are preserved very well with minimal unintended changes \\
4 & mostly preserved; minor unintended changes \\
3 & partly preserved; noticeable unintended changes \\
2 & poorly preserved; major unintended changes \\
1 & not preserved; extensive unintended changes \\
\end{tabular} \\[0.6em]

Output only one digit: 1, 2, 3, 4, or 5. No words, no punctuation, no explanation. \\
\hline
\end{tabular}
\end{table}

\begin{table}[h]
\centering
\caption{Prompt template for editing quality scoring.}
\label{tab:prompt_editing_quality}
\renewcommand{\arraystretch}{1.25}
\begin{tabular}{|p{0.95\linewidth}|}
\hline
\textbf{Prompt Template: Editing Quality Scoring} \\
\hline
You are an advanced vision–language model acting as a video quality rater. 
I will provide several sampled frames from a single edited video. \\[0.6em]

Assess the overall \textbf{visual quality} considering the following aspects:  
\begin{itemize}
    \item sharpness and detail preservation
    \item noise, compression artifacts, or blocking
    \item color banding and gradient smoothness
    \item flicker or temporal stability (as inferred from frames)
    \item exposure, contrast, and color accuracy
    \item deformations or obvious editing artifacts
\end{itemize} \\[0.6em]

Rate the overall visual quality on a 1–5 scale and output only one digit: \\[0.6em]

\begin{tabular}{@{}ll}
5 & excellent, very clear, minimal artifacts \\
4 & good, only minor artifacts \\
3 & fair, noticeable artifacts but acceptable \\
2 & poor, strong artifacts, blur, or flicker \\
1 & very poor, severe degradation \\
\end{tabular} \\[0.6em]

Output only one digit: 1, 2, 3, 4, or 5. No words, no punctuation, no explanation. \\
\hline
\end{tabular}
\end{table}

\begin{table}[t]
\centering
\caption{Prompt template for success ratio.}
\label{tab:prompt_success_ratio}
\renewcommand{\arraystretch}{1.25}
\begin{tabular}{|p{0.95\linewidth}|}
\hline
\textbf{Prompt Template: Success Ratio} \\
\hline
You are an advanced vision–language model tasked with verifying video edits. \\[0.6em]

The following images are sampled frames from the original and edited versions:  
the left sub-image shows the original, and the right sub-image shows the edited result.  
The two sub-images are separated by a blank divider. \\[0.6em]

The video-editing prompt for this case is: \verb|<instruct_prompt>| \\[0.6em]

Your task is to:  
\begin{enumerate}
    \item understand the content of the original video from its frames,
    \item examine the edited video frames and determine whether the changes match the editing prompt,
    \item ensure that no additional edits are introduced—only the modifications required by the prompt are allowed.
\end{enumerate} \\[0.6em]

You must reply with only a single lowercase word: \texttt{yes} or \texttt{no}.  
No explanations, punctuation, or extra text. \\
\hline
\end{tabular}
\end{table}
\end{document}